\documentclass[letterpaper]{article} 
\usepackage{aaai24}  
\usepackage{times}  
\usepackage{helvet}  
\usepackage{courier}  
\usepackage[hyphens]{url}  
\usepackage{graphicx} 
\usepackage{natbib}  
\usepackage{caption} 
\usepackage{xcolor}

\usepackage{newfloat}
\usepackage{listings}
\DeclareCaptionStyle{ruled}{labelfont=normalfont,labelsep=colon,strut=off} 
\usepackage[ruled,vlined]{algorithm2e}
\usepackage{caption}
\usepackage{subcaption}
\usepackage{graphicx}
\usepackage{booktabs}

\title{Cooperative Knowledge Distillation: A Learner Agnostic Approach}
\author {
    Michael Livanos,
    Ian Davidson,
    Stephen Wong
}
\affiliations {
    University of California, Davis\\
    mjlivanos@ucdavis.edu, lavidson@cs.ucdavis.edu, stswong@ucdavis.edu
}

\begin{document}

\maketitle

\begin{abstract}

Knowledge distillation is a simple but powerful way to transfer knowledge between a teacher model to a student model. Existing work suffers from at least one of the following key limitations in terms of direction and scope of transfer which restrict its use: all knowledge is transferred from teacher to student regardless of whether or not that knowledge is useful, the student is the only one learning in this exchange, and typically distillation transfers knowledge only from a single teacher to a single student. We formulate a novel form of knowledge distillation in which many models can act as both students and teachers which we call cooperative distillation. The models cooperate as follows: a model (the student) identifies specific deficiencies in it's performance and searches for another model (the teacher) who encodes learned knowledge into instructional virtual instances via counterfactual instance generation. Because different models may have different strengths and weaknesses, all models can act as either students or teachers (cooperation) when appropriate and only distill knowledge in areas specific to their strengths (focus). Since counterfactuals as a paradigm are not tied to any specific algorithm, we can use this method to distill knowledge between learners of different architectures, algorithms, and even feature spaces. We demonstrate that our approach not only outperforms baselines such as transfer learning, self-supervised learning, and multiple knowledge distillation algorithms on several datasets, but it can also be used in settings where the aforementioned techniques cannot.

\end{abstract}

\section{Introduction}
Knowledge distillation is a simple and elegant approach that allows one machine (the teacher) to instruct another machine (the student). Typically, the teacher model is more complex than the student model, and knowledge distillation compresses models for efficiency~\cite{hinton}, though more recent work explores improving performance as well~\cite{noisyStudent}. However, existing knowledge distillation has its limitations. First, offline knowledge distillation, that is, a trained teacher teaching an untrained student, assumes that \underline{all} of the teacher's knowledge is good and should be learned by the student even if the teacher performs worse than the student. Second, it is \underline{unidirectional} and \underline{singular}; one teacher informs one student, and students do not inform teachers.

In this work, we extend knowledge distillation to novel settings by creating what we call cooperative distillation. This is useful in domains where there are multiple learners, each of which can be considered a semi-expert deficient in one or more particular aspect(s) of a task, and can help overcome each other’s limitations. This setting is not covered by existing distillation work. Consider our FashionMNIST dataset experiment. Here, we create ten classifiers (one for each class) trained with one class being undersampled by 95\% to induce a conceptual deficiency. A model might understand the majority of clothes it sees, but since it hasn't seen many, say, ankle boots, it struggles to classify them correctly and will rely on other models to teach it this concept. This will require targeted and multidirectional transfer: this model needs to be taught only about ankle boots and can be a teacher for other classes. 

In the tradition of knowledge distillation simplicity, we propose a learner agnostic, counterfactual-based cooperative approach. Consider an instance $x$ which model $i$ can predict correctly, but model $j$ cannot. We say that model $i$ is a qualified teacher to model $j$ for the specific instance $x$. Our method will have model $i$ teach model $j$ about $x$ by generating a new type of \underline{quintessential} counterfactual $x'$ which can be added to $ j$'s training set. We call this type of counterfactual quintessential because instead of modifying the instance to change its label, we have the model $i$ make this instance look even more like the true class. Counterfactuals were chosen as the method to generate virtual instances since they are both model agnostic and virtual instance generation is driven by the model. Our approach is multidirectional as any model can teach any other and focused as we transfer only some instances between models via counterfactuals.

Our work can be viewed as being in a similar setting to domain adaptation and transfer learning but has notable differences. Typically, domain adaptation is from a \underline{chosen} single expert source to a single novice target, whereas our work is cooperative between semi-experts with no need to choose a target/source. Further in our work, the domain of the teacher and student models are the same which is not the case for transfer learning.
Our contributions are:

\begin{itemize}
    \item \emph{New Style of Distillation.} We propose a simple yet powerful approach to a new form of distillation we call cooperative distillation. This is achieved using a novel type (quintessential) and use of counterfactuals.
    \item \emph{Robust Across Learners.} Experimental results are promising for a variety of basic (i.e., decision trees) and complex learners (i.e.,  convolutional neural networks) (see Experimental Section, particularly Table \ref{tab:resultTable}).
    \item \emph{Robust Across Settings.} We demonstrate our method's good performance under various settings, including distilling between different architectures/algorithms, high-performance models, low-performance models, mixtures of high and low-performance models and varying degrees of feature overlap.
    \item \emph{Outperforms Baselines.} Our approach can significantly outperform multiple state-of-the-art and state-of-the-practice baselines in transfer learning, self-supervised learning, and knowledge distillation.
    (see Table \ref{tab:resultTable} which summarizes all our experiments).
\end{itemize}

We begin this paper by outlining related work and describing our approach. We then provide experimental results for various learners, followed by a discussion on our method's strengths and weaknesses including our hypotheses why our 
 method works, after which we conclude.

\section{Related Work}

The field of knowledge distillation exists to transfer learned information from one learner to another, typically a more costly high-performance model to a lightweight model~\cite{hinton}\cite{bertKD}\cite{survey} in the \underline{same task}. This is distinct from transfer learning which, by definition, uses a learner in a \underline{related but different} source domain to assist in the training of a learner in the target domain \cite{zhuang2020comprehensive}. Our work is further differentiated by distilling knowledge between semi-experts in a multidirectional fashion, as opposed to an expert to a novice.

Knowledge distillation literature can be categorized by two main factors: what is considered knowledge and the distillation scheme~\cite{survey}. We first discuss how these questions have been answered by previous work and then present our novel knowledge paradigm.

\noindent
\textbf{Knowledge Distillation Paradigms.}
There are three general categories of knowledge distillation algorithms: \underline{response/output distillation} in which a student learns to replicate the output of the learner by calculating loss between the student's logits and those of the teacher~\cite{hinton}~\cite{response2}, \underline{feature distillation} which trains a student to mimic the teacher's parameters, such as hidden layer weights and biases,~\cite{feature2}~\cite{feature3}~\cite{online_feature1}, and \underline{relation distillation}, which is concerned with the relations between multiple parts of the model such as multiple feature maps~\cite{crosslayer}~\cite{betweenFeatures}, feature maps and logits~\cite{instanceRelation}, or pairwise similarities between the input data and output distribution~\cite{pairwise}.

Distillation schemes are also important to categorize the different forms of knowledge distillation. Knowledge can be distilled from a learned teacher model to a student in offline knowledge distillation~\cite{hinton}~\cite{bertKD}~\cite{betweenFeatures}, or while the learned model is being trained in online distillation~\cite{online1}~\cite{coDist}. It is important to note that some of these approaches do consider multiple learners both students and teachers~\cite{mutual} (one of our contributions), however, the task for those approaches is to distill knowledge during the lurking process, whereas we are distilling knowledge between trained models, which is novel for this setting.

\begin{figure*}
    \centering
    \includegraphics[scale=0.36]{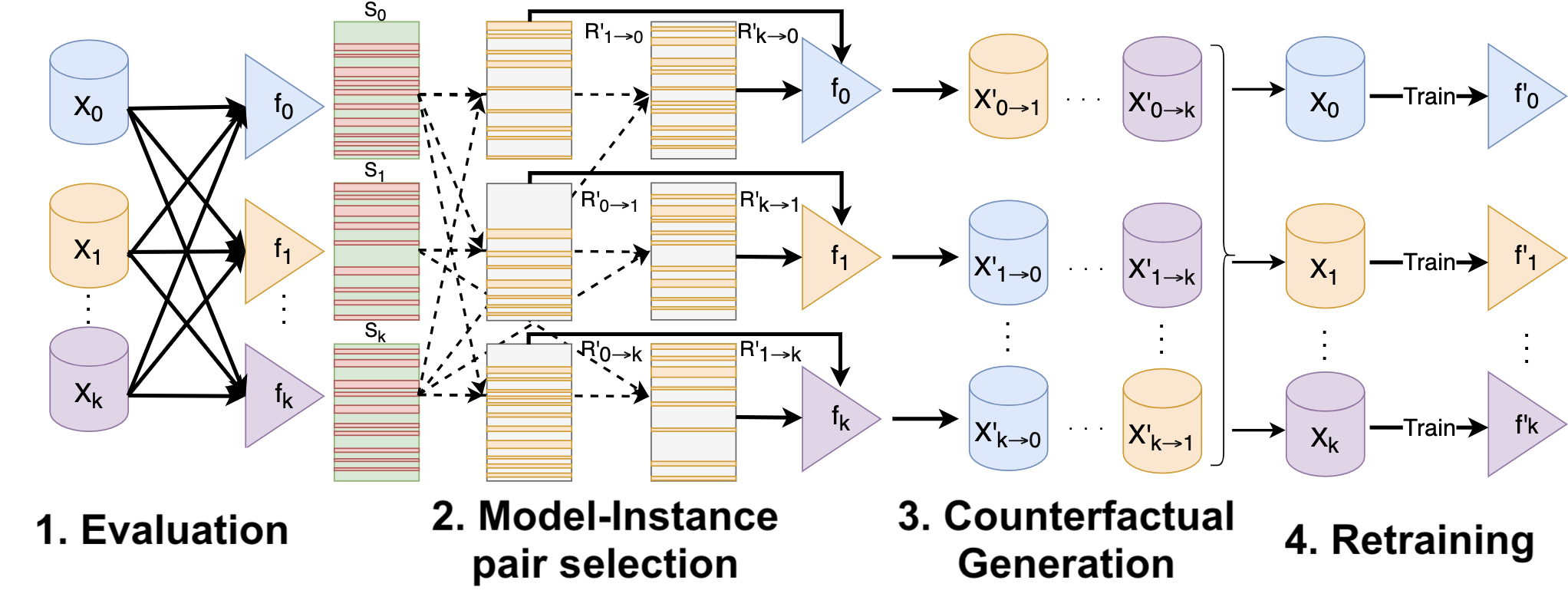}
    \caption{Pipeline for our method. Each model and it's corresponding datasets are color coded. Given $k + 1$ models $f_0$ through $f_k$ and datasets $X_0$ through $X_k$, we find which instances among the training sets each model is able to correctly predict (rows colored in green), creating a set of indices for each model (step 1). For all permutations of groups of two of these sets ($S_i$, $S_j$), we find $R_{i \rightarrow j} = S_i - S_j$: instances for which model $i$ is a teacher for model $j$ (rows highlighted in yellow, step 2). We then create counterfactuals using the appropriate teacher models and instances, labeled $X'_{teacher \rightarrow student}$ (step 3), shuffle the new, augmented instances into the training sets of each group, and retrain the models to create augmented models $f'_0$ to $f'_k$ (step 4).}
    \label{fig:approach}
\end{figure*}

\noindent
\textbf{A New Use for Counterfactuals - Cooperative Distillation} 
Rather than encoding knowledge into output logits, parameters, or relationships, our work embeds learned information in the data itself by creating virtual instances (counterfactuals) and passing them on to the training sets of other models. Further, it should be noted that the distillation scheme is also a special case of offline knowledge distillation, as instead of a student learning from a teacher, each model will act as both teacher and student simultaneously, something not explored in offline knowledge distillation. This is distinct from both self-distillation~\cite{selfDist}, in which a single model acts as both teacher and student, as our method uses multiple models, and online distillation\cite{online1} in which models distill knowledge during training, as our method leverages trained models.

\section{Our Approach: Cooperative Distillation}


Our method is a form of offline knowledge distillation, but with two important enhancements. First, it considers distillation across \underline{multiple} models where each model can act as both a teacher and student, rather than distilling from a single teacher to a student. Our second innovation uses counterfactuals to generate \underline{targeted} instances to transfer rather than distilling knowledge across all instances as in traditional knowledge distillation. This is a form of cooperation as the student identifies instance it performs poorly on and the teacher creates an easier to understand counterfactual.

Our approach takes three fundamental steps:
\begin{enumerate}
\item
\textbf{Expertise Identification:} Model $i$ selects instances ($I$) it can accurately predict. 
\item 
\textbf{Deficiency Identification:} From $I$, every other model $j$ finds instances it cannot predict $R_{i \rightarrow j} \subset I$.
\item
\textbf{Cooperative Distillation:} For each instance $x \in R_{i \rightarrow j}$, $i$ creates counterfactual $x'$ to be added to $j$'s training set.
\end{enumerate}

\subsection{Expertise and Deficiency Identification} Since each model may have limited knowledge of the domain, it is crucial that models acting as teachers only do so in settings where they are "qualified" teachers. A model $i$ is considered qualified to teach a student model $j$ about an instance $x$ if and only if model $i$ correctly predicts instance $x$ and model $j$ does not. In this way, students are only taught concepts they fail to understand and only from  qualified teachers.

\begin{figure}
    \centering
    \includegraphics[scale=0.28]{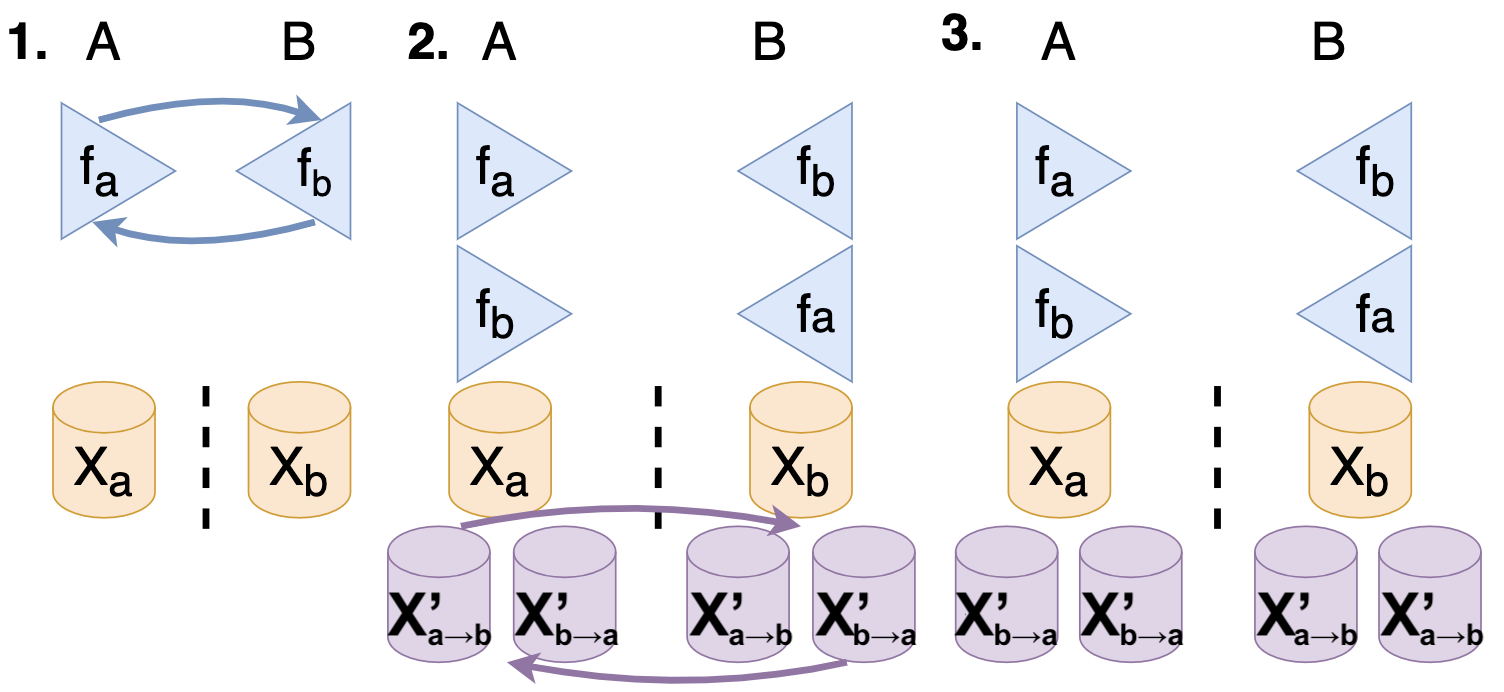}
    \caption{Non-Data Sharing Scenario: Our approach for deploying the approach while maintaining data privacy. Institutions may share models, but not data. Models are exchanged (step 1), our technique is applied to create a subset of the virtual instances (step 2), those virtual instances are shared, and the models are retrained (step 3).}
    \label{fig:private}
\end{figure}

To decide which models act as students and which act as teachers for different instances, we first pass all of the training data $X$ (this can be done without sharing data, see Figure \ref{fig:private}) to each of the models and collect sets of indices of the instances that model can predict correctly. Let $S_i$ be the set of instance indices correctly predicted by model $i$. Let $R_{i \rightarrow j}$ be the set of instance indices that model $i$ correctly predicts that $j$ does not.
Formally:

\begin{equation}
\label{eq1}
    R_{i \rightarrow j} = S_i - S_j
\end{equation}

This can be accomplished even if datasets cannot be shared. Consider two groups/organizations/sites who can share models, but not data. After sharing models, they can use our approach to generate virtual instances on their respective datasets and only share those virtual instances. This process can be visualized in Figure \ref{fig:private}. Equation \ref{eq1} must be computed for every permutation of two models. Therefore for $k$ models, the complexity of this subroutine is $O(P^k_2 |X|)$, where $|X|$ is the size of the training data.

\subsection{Quintessential Counterfactual Generation}

Counterfactual algorithms generate a virtual instance $x'$ given three pieces of information: the model $f$, an instance $x$, and desired output $y'$ such that $x'$ is similar to $x$ and $f(x') = y'$~\cite{CFSurvey}. Most work creates \underline{contrastive} counterfactuals which ``flip" the label, ($f(x) \neq f(x')$) whereas our method generates quintessential counterfactuals - those which the existing prediction is made greater.

\begin{figure}
    \centering
    \includegraphics[scale=0.17]{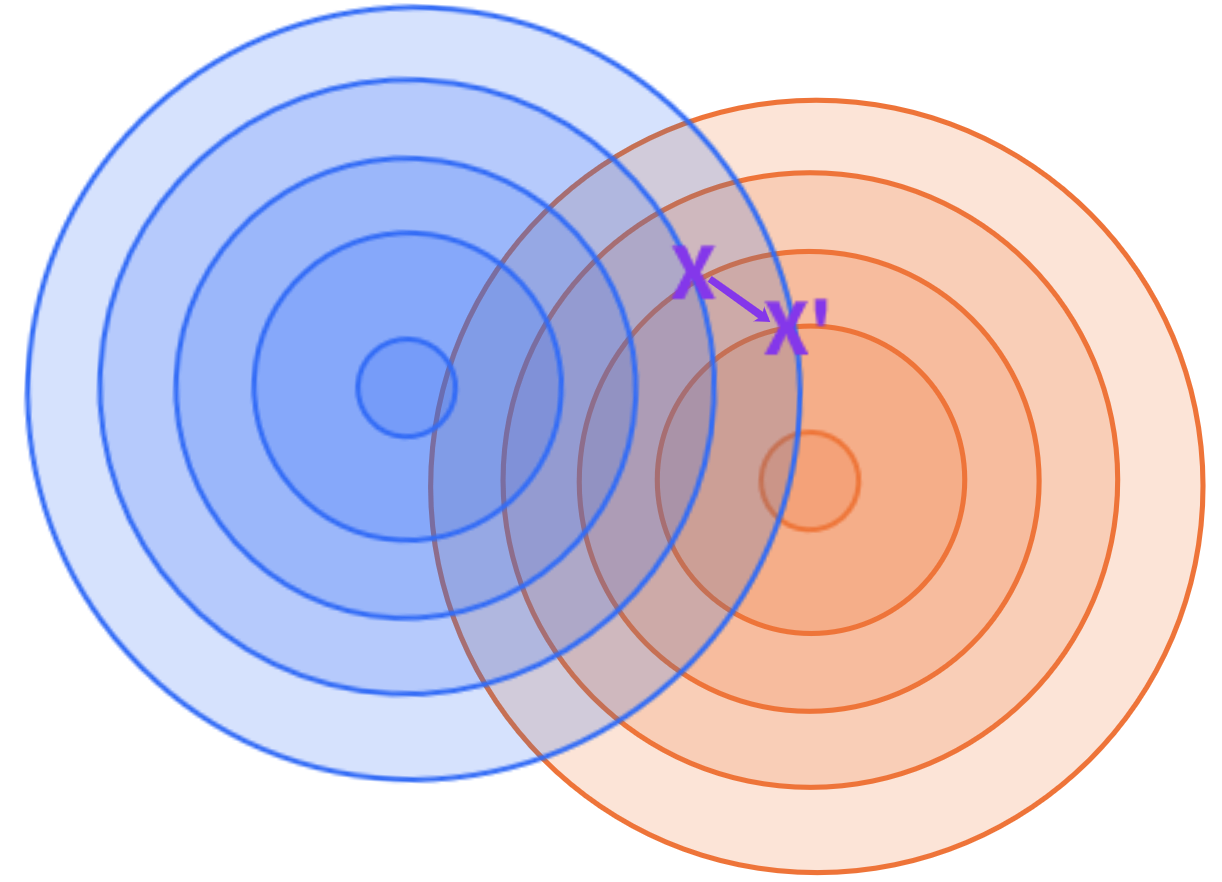}
    \caption{Quintessential counterfactual generation illustrative example. Each model's decision surface is a contour map with each circle representing 20\% confidence of the correct class. The orange model (teacher) predicts instance $x$ with 60\% confidence, and the blue model (student) mispredicts $x$ as it's confidence in the correct class is only 40\%. 
    The teacher model creates a virtual instance $x'$ it believes is more typical of the class.}
    \label{fig:cfgen}
\end{figure}

The instance selection mechanism previously described finds the appropriate instance-teacher pair ($f,x$). We chose to set $y'$ = $f_i(x) + \alpha(y - f_i(x))$, where $\alpha$ encodes the  teacher model's influence. The closer $\alpha$ is to 1 the closer $y'$ is to $y$ (the true class label), and the closer $\alpha$ is to 0, the closer $y'$ is to $f_i(x)$. This allows the teacher model to inject knowledge about the class into the instances. All experiments in this paper set $\alpha$ to 0.5, as this is an even balance between the original instance and a theoretical instance of perfect certainty. Conceptually, our counterfactual generation process can be visualized in Figure \ref{fig:cfgen}.

The counterfactuals are assigned the correct label for the original instance ($y$), the augmented instances are added to the training set of the student, and the model is retrained. This process is described in Figure \ref{fig:approach} and Algorithm 1. Counterfactuals for model $j$ from model $i$ are generated as below:

\begin{equation}
\label{eq2}
    \forall x \in R_{i\rightarrow j}~argmin_{x'}\: d(x', x) + \lambda|f_j(x') - y'|^2
\end{equation}
where $R_{i \rightarrow j}$ are the instances model $i$ can teach model $j$.

Here, $d$ is a distance metric, in our case Manhattan distance, and $\lambda$ is a balance term. As is standard\cite{molnar2019}, we set $\lambda$ to the maximum value for which a solution will converge. For models with differentiable parameters, such as neural networks, we use gradient descent via Adam, and for models without differentiable parameters, the non-gradient-based particle swarm optimization~\cite{pso}. The complexity to generate a counterfactual is constant, making the cost $O(k|X|)$ for k models.

Assuming the cost to train $k$ models is proportional to the data size, the algorithm's time complexity is $O(P^k_2 |X| + k |X|)$, or with a constant number of learners $O(|X|)$.

\begin{algorithm}
    \SetKwInOut{Input}{input}
    \SetKwInOut{Output}{output}

    \Input{$k+1$ trained models $F={f_0,f_1,...f_k}$, respective datasets $D=(X_0 , Y_0),(X_1, Y_1),...(X_k , Y_k)$ and balance term $\alpha$.}
    \Output{$k$ retrained models}
    \tcp{Expertise Identification}
 $S$ := new list of sets\;
 \For{$(X_i,y_i) \in D$}{
    \For{$f \in F$}{
        $S_i = f.scorePredictions(X_i, y_i)$
    }
 }
 \tcp{Counterfactual Generation}
 $AugmentedInstances$ := new list of instances\;
 \For{$\forall (S_i, S_j) \in S$}{
    $R_i$ := $S_i - S_j$ (Eq. \ref{eq1})\;
    \For{$(x,y) \in D[R_i]$}{
        $y'$ := $f_i(x) + \alpha * (y - f_i(x))$\;
        $cf$ := $GenerateCFs(f_i,x,y')$ (Eq. \ref{eq2})\;
        $AugmentedInstances[j].append(cf)$\;
    }
 }
 \tcp{Model Retraining}
 \For{$i = 0 \rightarrow k$}{
    $newDataset$ := $AugmentedInstances[i] + D_i$ \;
    $f_i.train(newDataset)$ \;
 }
 \caption{Cooperative knowledge distillation.}
\end{algorithm}

\subsection{Extensions for Mismatched Feature Sets}
\label{sec:mismatch}
Some practical situations exist where the feature sets are not identical but overlap. This can occur for a variety of reasons including if data is collected from different locations or sites. All models are tested on the same test set, which also comes from a different dataset from another site containing no instances from any training or validation sets. Consequently, some datasets may have different features, and we must therefore pass instances into training sets of incompatible feature spaces. To deal with this, we normalize continuous ratio and interval data between zero and one, and one-hot encode categorical and discrete interval data, setting missing features to zero, as suggested in ~\cite{delMissing}\cite{replaceZero}. 

\section{Experiments\footnote{To aid in reproducibility, code is provided on GitHub \url{https://github.com/MLivanos/Cooperative-Knowledge-Distillation}}}
\label{sec:exp}

\begin{table*}
\resizebox{\textwidth}{!}{%
\begin{tabular}{@{}llllllllll@{}}
\toprule
\textbf{}                                   & \textbf{}         & \multicolumn{1}{l|}{\textbf{}}     & \multicolumn{1}{l|}{\textit{\textbf{\begin{tabular}[c]{@{}l@{}}Transfer\\ Learning\end{tabular}}}} & \textit{\textbf{SSL}} & \multicolumn{1}{l|}{\textit{\textbf{}}} & \textit{\textbf{Knowledge}}                                           & \textit{\textbf{Distillation}}                                       & \multicolumn{1}{l|}{\textbf{}}                                                             & \textit{\textbf{\begin{tabular}[c]{@{}l@{}}Data-\\ Pollution\end{tabular}}}   \\ \midrule
\textbf{Approach\textbackslash Experiment} & \textbf{Baseline} & \multicolumn{1}{l|}{\textbf{Ours}} & \multicolumn{1}{l|}{\textbf{\begin{tabular}[c]{@{}l@{}}Parameter\\ Transfer\end{tabular}}}         & \textbf{GAN}                      & \multicolumn{1}{l|}{\textbf{Mixup}}             & \textbf{\begin{tabular}[c]{@{}l@{}}Response-\\ Based KD\end{tabular}} & \textbf{\begin{tabular}[c]{@{}l@{}}KD as\\ Pretraining\end{tabular}} & \multicolumn{1}{l|}{\textbf{\begin{tabular}[c]{@{}l@{}}Parameter\\ Based KD\end{tabular}}} & \textbf{\begin{tabular}[c]{@{}l@{}}Add Training\\ Data Together\end{tabular}} \\ \midrule
\textbf{Exp. 1 CL MLP}                & 60.98\%           & \textbf{68.68\%}                   & N/A                                                                                                & 65.13\%                           & 52.56\%                                         & 59.29\%                                                               & 52.43\%                                                              & 61.24\%                                                                                    & 67.01\%                                                                       \\ \midrule
\textbf{Exp. 1 AE MLP}                & 86.04\%           & \textbf{87.74\%}                   & N/A                                                                                                & 86.58\%                           & 62.71\%                                         & 60.00\%                                                               & 85.87\%                                                              & N/A                                                                                        & 84.43\%                                                                       \\ \midrule
\textbf{Exp. 2 CL 1 D-Tree}           & 63.41\%           & \textbf{67.58\%}                   & N/A                                                                                                & 62.41\%                           & 55.63\%                                         & 58.99\%                                                               & N/A                                                                  & 62.53\%                                                                                    & 69.29\%                                                                       \\ \midrule
\textbf{Exp. 2 AE MLP}                & 86.04\%           & \textbf{87.32\%}                   & N/A                                                                                                & 86.58\%                           & 62.71\%                                         & 58.03\%                                                               & 86.72\%                                                              & N/A                                                                                        & 84.43\%                                                                       \\ \midrule
\textbf{Exp. 3 Model 1 D-Tree}        & 56.71\%           & \textbf{57.44\%}                   & N/A                                                                                                & 56.38\%                           & 53.38\%                                         & 54.55\%                                                               & N/A                                                                  & N/A                                                                                        & 60.72\%                                                                       \\ \midrule
\textbf{Exp. 3 Model 2 D-Tree}        & 43.45\%           & \textbf{62.36\%}                   & N/A                                                                                                & 51.49\%                           & 57.16\%                                         & 55.45\%                                                               & N/A                                                                  & N/A                                                                                        & 62.54\%                                                                       \\ \midrule
\textbf{Exp. 3 Model 3 D-Tree}        & 53.47\%           & \textbf{63.04\%}                   & N/A                                                                                                & 55.45\%                           & 58.22\%                                         & 62.27\%                                                               & N/A                                                                  & N/A                                                                                        & 55.17\%                                                                       \\ \midrule
\textbf{Exp. 4 Model 1 D-Tree}        & 56.71\%           & \textbf{66.24\%}                   & N/A                                                                                                & 56.38\%                           & 60.89\%                                         & 54.55\%                                                               & N/A                                                                  & N/A                                                                                        & 60.72\%                                                                       \\ \midrule
\textbf{Exp. 4 Model 2 NB}            & 62.37\%           & \textbf{77.56\%}                   & N/A                                                                                                & 64.69\%                           & 70.41\%                                         & 54.13\%                                                               & N/A                                                                  & N/A                                                                                        & 64.03\%                                                                       \\ \midrule
\textbf{Exp. 4 Model 3 SVM}           & 54.45\%           & \textbf{59.08\%}                   & N/A                                                                                                & 55.78\%                           & 52.96\%                                         & 54.13\%                                                               & N/A                                                                  & N/A                                                                                        & 58.42\%                                                                       \\ \midrule
\textbf{Exp. 5 Median CNN}            & 76\%              & 83\%                               & 82\%                                                                                               & 81\%                              & 79\%                                            & 73\%                                                                  & \textbf{86\%}                                                        & 80\%                                                                                       & 86\%                                                                          \\ \bottomrule
\end{tabular}
}
\caption{Median results from 10 to 90 experiments. Methods that cannot be used for a particular dataset are marked with N/A. In all four of the main experiments (1-4) our method outperforms all baselines and competitors. The stress test in Experiment 5 designed to test our method’s ability to handle many models achieves the second highest performance, with knowledge distillation as pertaining performing best. The baselines are models are trained without any augmentation.}
\label{tab:resultTable}
\end{table*}

To demonstrate our claims  we conduct six experiments, using models generated from five different algorithms trained on nine datasets for four tasks.
Recall that our claim was that our model agnostic cooperative distillation involves focused and multi-way distillation. Each experiment is meant to examine one particular aspect of these claims and compare them to other relevant state-of-the-art and state-of-the-practice methods. We next  discuss the implications of each experiment  and provide detail in subsequent subsections.

\begin{figure*}[ht]
    \centering
    \begin{subfigure}{0.31\textwidth}
        \centering
        \includegraphics[scale = 0.16]{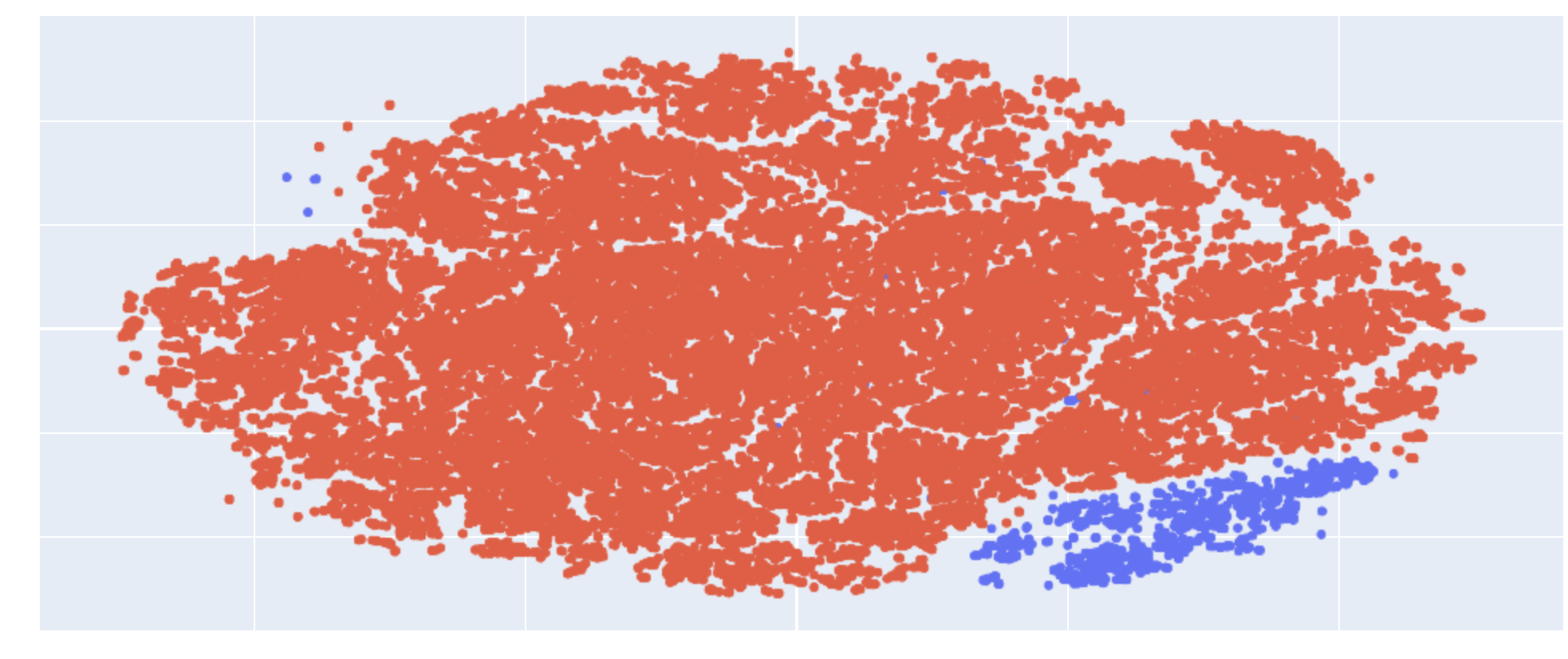}
        \caption{Distribution of original datasets for datasets 1 and 2}
    \end{subfigure}
    \begin{subfigure}{0.31\textwidth}
        \centering
        \includegraphics[scale = 0.16]{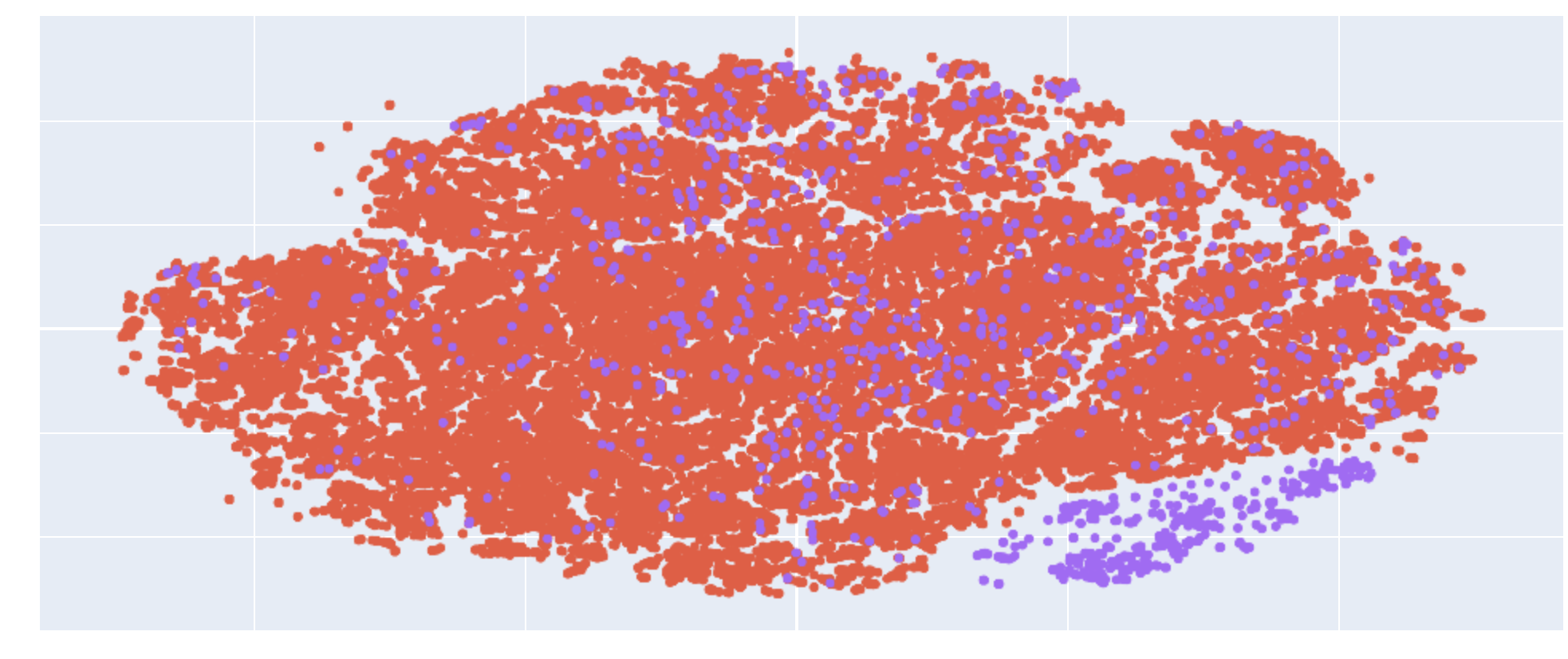}
        \caption{Original and augmented instances for dataset 1}
        \label{fig:mode1arg}
    \end{subfigure}
    \begin{subfigure}{0.31\textwidth}
        \centering
        \includegraphics[scale = 0.16]{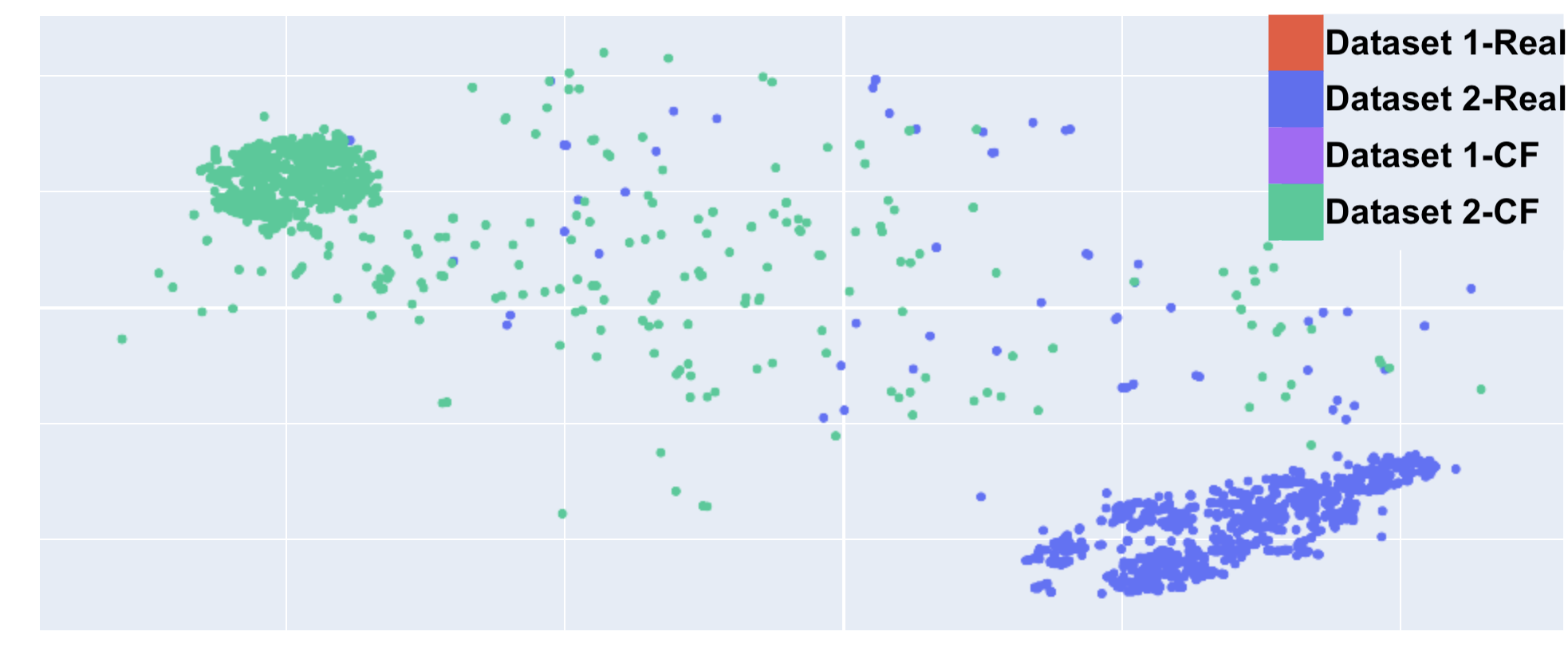}
        \caption{Original and augmented instances for dataset 2}
        \label{fig:mode2arg}
    \end{subfigure}
    \begin{subfigure}{0.48\textwidth}
        \centering
        \includegraphics[scale = 0.175]{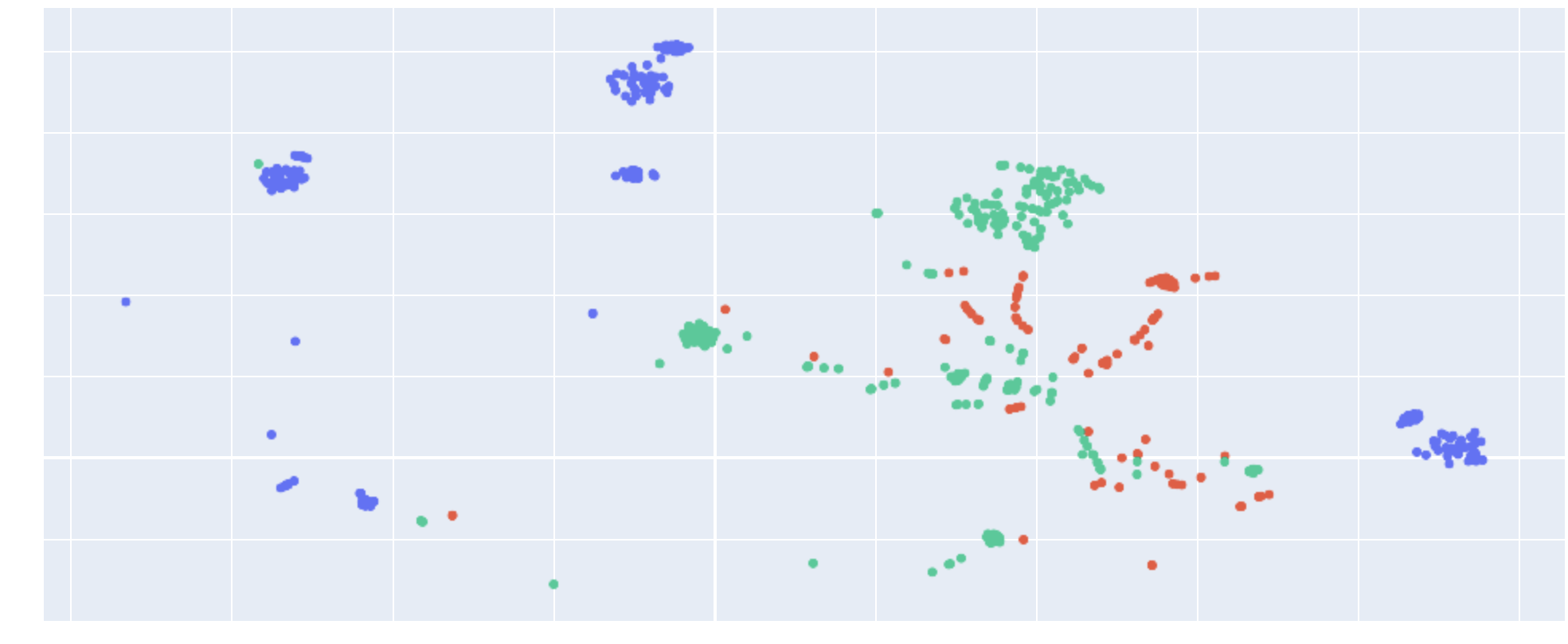}
        \caption{Distribution of original datasets}
    \end{subfigure}
    \centering
    \begin{subfigure}{0.48\textwidth}
        \centering
        \includegraphics[scale = 0.175]{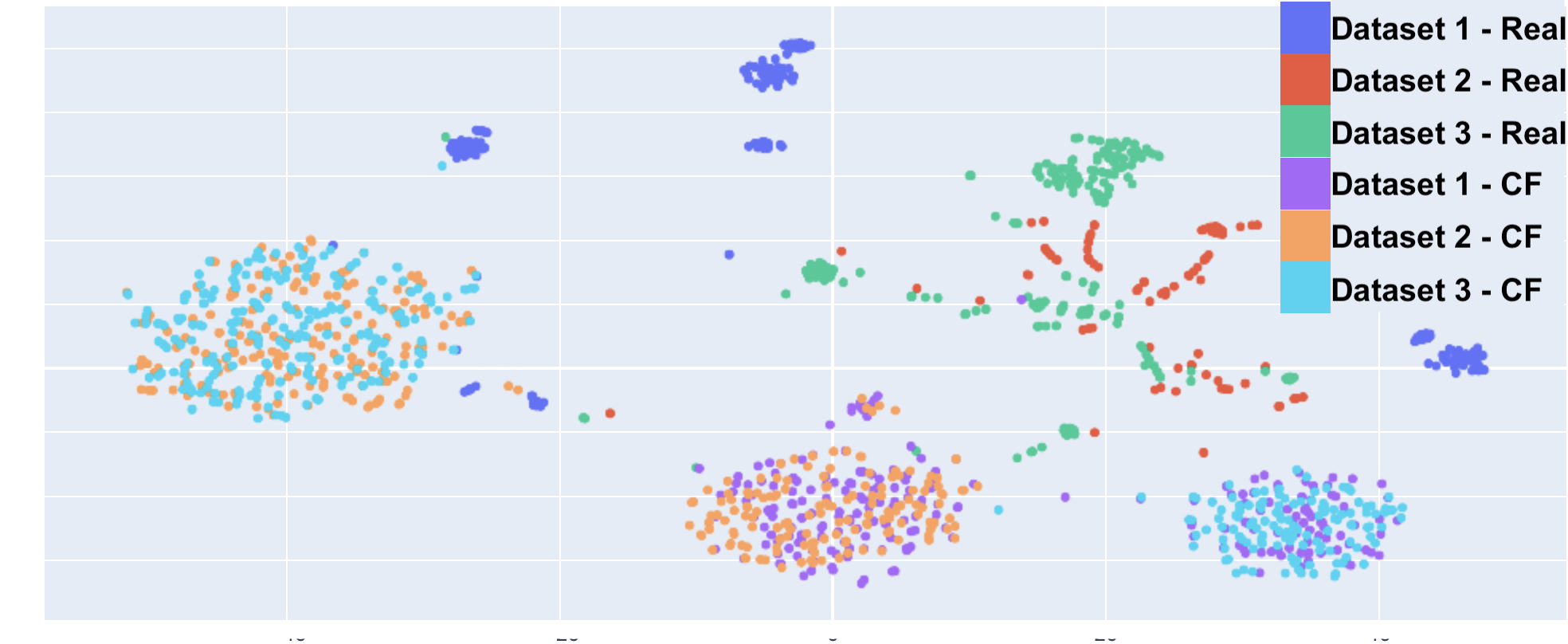}
        \caption{Distribution of original datasets and added augmented instances}
    \end{subfigure}
    \caption{All data points, original and augmented, for both datasets of Experiment 1 (top; subfigures a-c) and Experiment 4 (bottom; subfigures d,e), projected into two dimensions using 
    t-SNE. Each model is able to create instances similar to their own data while being distinct from the original data points.}
    \label{fig:exp1Distrib}
\end{figure*}

\begin{figure}
    \centering
    \includegraphics[scale=0.225]{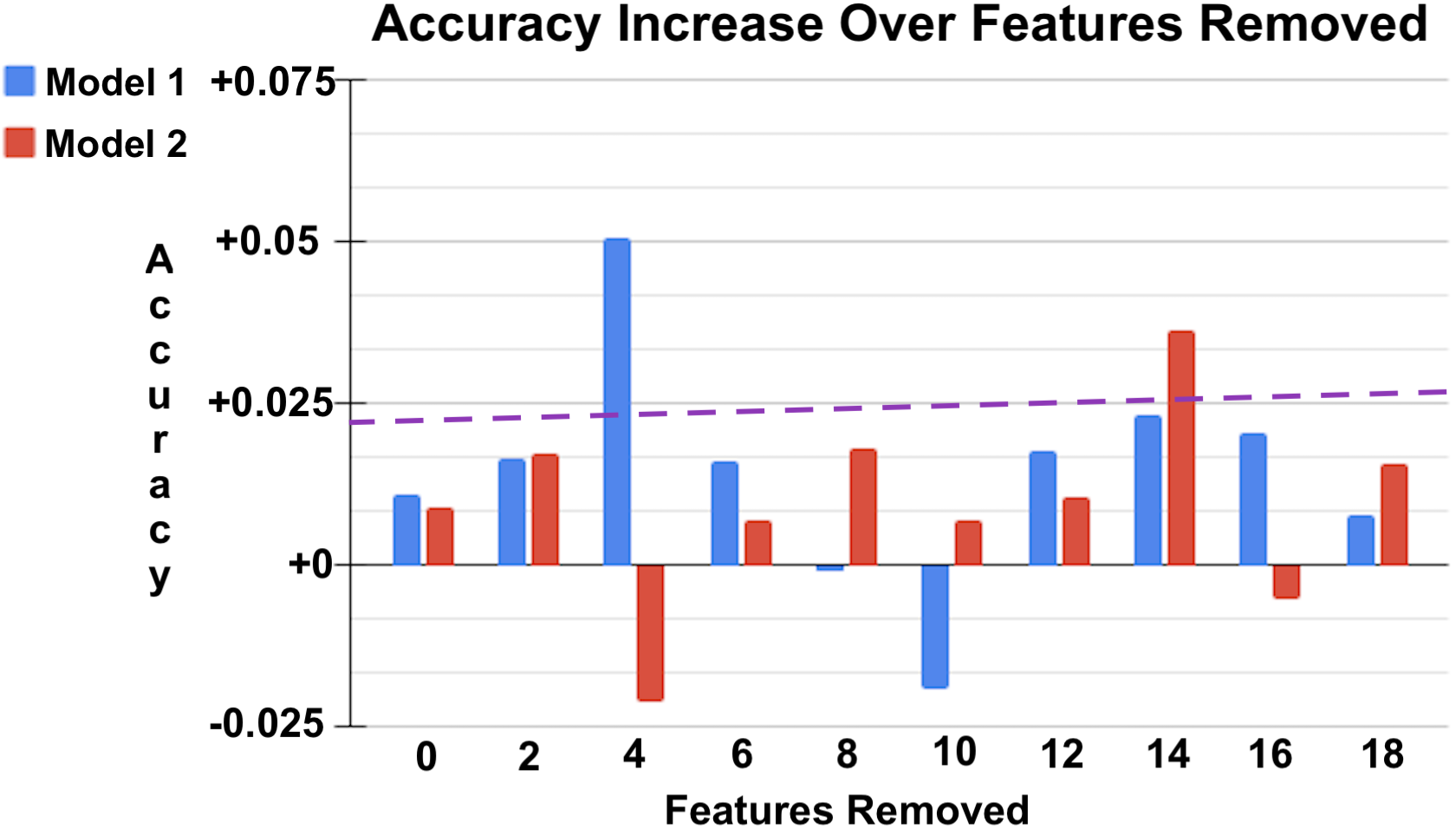}
    \caption{Improvement in accuracy over the number of features removed (feature overlap becomes smaller along the X-axis). There is no significant correlation between feature overlap and performance gain, as demonstrated by the trendlines which average a slope of $2.25*10^{-4}$.}
    \label{fig:AccOverOverlap}
\end{figure}

\begin{itemize}
    \item Experiments 1 and 2 examine how our approach handles distilling between different architectures (Experiment 1) and algorithms (Experiment 2), as well as differing amounts of data and performance. This asymmetrical setting leads to different numbers of counterfactuals generated for each model (see Figure \ref{fig:exp1Distrib}).
    \item Experiments 3 and 4 use three models to test our model's multidirectional claim as each model has a small amount of training data and all need to cooperate to master the domain. Further, these models start at a relatively weak performance, meaning we are also testing our focus mechanism to ensure that only relevant knowledge is distilled. Experiment 4 pushes the limits on our model-agnostic claim as we have a decision tree, Naive Bayes classifier, and SVM cooperating via our method.
    \item Experiment 5 tests many aspects of our claims at once: the ability to distill between many semi-expert models doing focused transfer. We create the situation where the training data is made deficient in exactly one class for each of the ten convolutional neural networks.
    \item Our last experiment  addresses our claim that our method can be used in settings with different amounts of overlapping features by starting with perfect feature overlap and iteratively removing features to test correlation between feature overlap and accuracy increase. See Figure \ref{fig:AccOverOverlap}.
    \item Notably, 
 with the exception of Experiment 5, no two datasets have a single instance in common with each other, instead relying on the process outlined in Figure \ref{fig:private} to accomplish our technique without sharing data.
\end{itemize}

The results of our experiments are summarized in Table \ref{tab:resultTable}. An important aspect of the results is that our method improves all 20 models trained, which did not occur with any competitor. The rest of this section will discuss the results and implications of each experiment individually.

All models discussed trained to convergence, and hyperparameter selection maximized validation set accuracy.

\noindent
\textbf{Baselines and Competitors.}
Each experiment tests against several competitors: parameter transfer\cite{parameterTransfer}, self-supervised learning techniques including generative adversarial networks (Deep Convolutional GAN (DCGAN)~\cite{dcgan} for image datasets and TabGan~\cite{tabgan} for tabular datasets) and mixup\cite{mixup}, and knowledge distillation, including response-based offline knowledge distillation~\cite{hinton}, response-based knowledge distillation which achieved state-of-the-art accuracy on the Imagenet dataset~\cite{noisyStudent}, and finally a recent, state-of-the-art feature-based knowledge distillation~\cite{response1} algorithm. We also compare against a baseline of the original model's accuracy (without distillation) and an \underline{idealized} setting in which all training data is combined. This last setting may be unrealistic due to data proprietary, privacy, or availability and is thus compared separately.

\noindent
\textbf{Experiments 1 \& 2: Cross-Architecture/Algorithm Distillation.}
These experiments use three different datasets to predict if a used car is expensive ($>$\$20,000) or inexpensive ($\leq$\$20,000). Each dataset comes from a different website curated between 2020 and 2021. Datasets 1 and 2 come from Craigslist~\cite{craigslistData} and Auction Export~\cite{AEData} respectively, and are used for training and validating models. A test set from Car Guru~\cite{CGData} simulates a future distribution all models will have to predict. We expect that each data set covers different types of cars (eg makes, models, years) in different depths.

Experiment 1 examines how our technique can distill knowledge between models of different architectures which were tuned for the different data sets. The Craigstlist (CL) and Auction Export (AucEx) models use neural networks of different \underline{architectures}: the former with one hidden layer with 512 neurons, the latter with one hidden layer with 1024 neurons, both with leaky ReLU activation functions for hidden layers and sigmoid for the output layer. These architectures create models with test set accuracies of 60.98\% and 86.04\% for the CL and AucEx models respectively, and our method improves this to 68.68\% and 87.74\%. These results not only demonstrate a boost to both model's performances but also show that a \underline{low-performance} model can teach an \underline{high-performance} model - a result that no other distillation technique could replicate. In the case of the AucEx model, our method performed even better than the idealized case of training using all available instances. A total of 18923 instances were distilled to the CL model and 1613 to AucEx.

In Experiment 2, the AucEx model is the same neural network, and the CL model is now a \underline{decision tree} (minimum samples leaf set to 20) to explore how well knowledge can be distilled between different \underline{algorithms}. The AucEx model's baseline performance is identical to the above experiment, and the CL model achieves baseline test accuracy of 63.41\%. Our method successfully elevates performance of the models to 67.01\% and 87.32\% for the CL and AucEx models, respectively, again surpassing all competitors. A total of 30842 instances were distilled to the CL model and 1817 to the AucEx model.

\noindent
\textbf{Experiment 3 \& 4: Small Data Distillation.}
Experiment 3 tests how well knowledge can be distilled between three low-performance models built from small datasets. Four datasets are used, each predicting the presence or absence of heart disease from hospitals at different locations: Long Beach (Model 1), Switzerland (Model 2), Hungary (Model 3), and Cleveland~\cite{Heart}, all sourced from \cite{uci}. The Cleveland was chosen as the test set since it contained all of the features of the previous three, making evaluation fairer. We remove features from each dataset individually if at least 25\% of instances are not reported and train decision trees for each dataset. Baseline test set accuracies for each model are 56.71\%, 43.45\%, and 53.47\%, which are improved to 60.72\%, 62.54\%, and 55.17\%, respectively by our method, greater than all applicable competitors, and in the third case, beating the idealized scenario of adding all instances. This demonstrates our method's ability to distill knowledge even between low performance learners.

Experiment 4 uses the same datasets, however instead of using models from the same algorithm, knowledge will be distilled between several different algorithms: a decision tree, naive Bayes, and support vector machine classifier. Model 1 is the same decision tree as in experiment 3, and models 2 and 3's baseline performances stand at 62.37\%, and 54.45\%, respectively. With comparably stronger baselines, our method elevates performance to 66.24\%, 77.56\%, and 59.08\%, surpassing all competitors and the idealized scenario of pooling together all instances. This experiment provides further evidence to suggest that our method not only can handle distilling  between  different  algorithms  but  is  largely invariant to algorithm choice and that stronger teacher models tend to help more than weaker ones.

\begin{figure}
    \centering
    \includegraphics[scale=0.18]{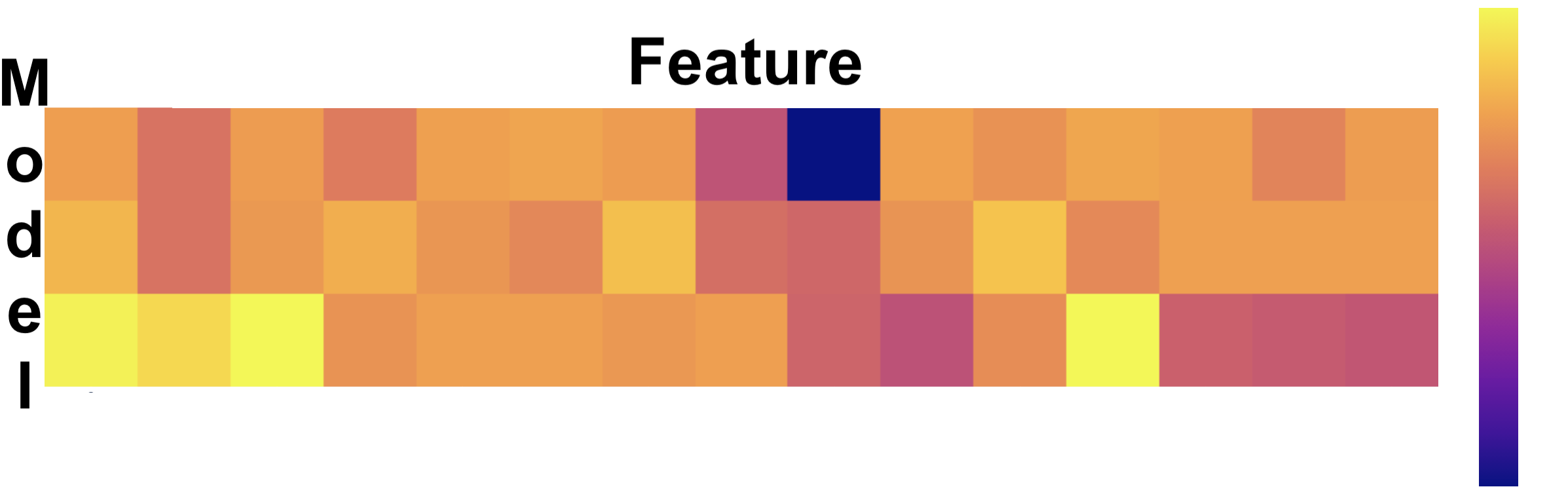}
    \caption{Heatmap for the knowledge distilled into different models of Experiment 4. Columns are features, and the rows represent the average change (yellow is positive, blue is negative) to move an instance to the diseased class. For example, the bottom left tile shows an increase in age (first column) is associated with heart disease whilst the middle top row indicates a reduction in ST-Depression (ninth row) decreases heart disease.
    }
    \label{fig:exp4Heatmap}
\end{figure}

\begin{figure}
    \centering
    \begin{subfigure}{0.19\textwidth}
        \includegraphics[scale=0.13]{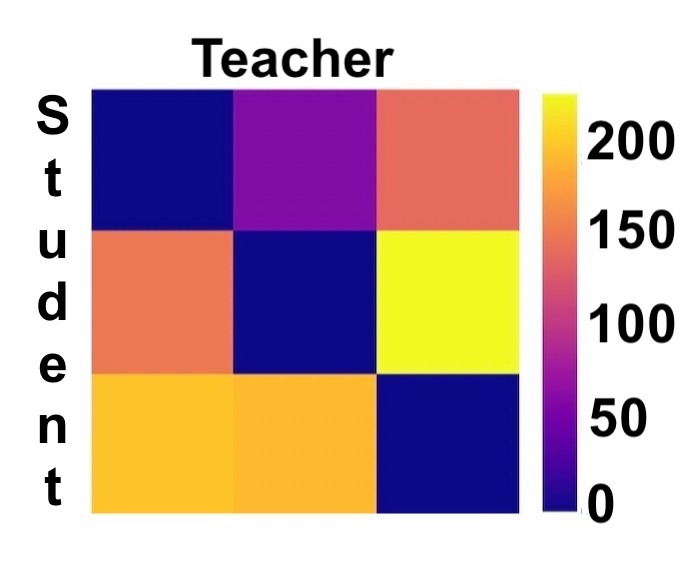}
        \caption{Experiment 3}
    \end{subfigure}
    \begin{subfigure}{0.19\textwidth}
        \includegraphics[scale=0.13]{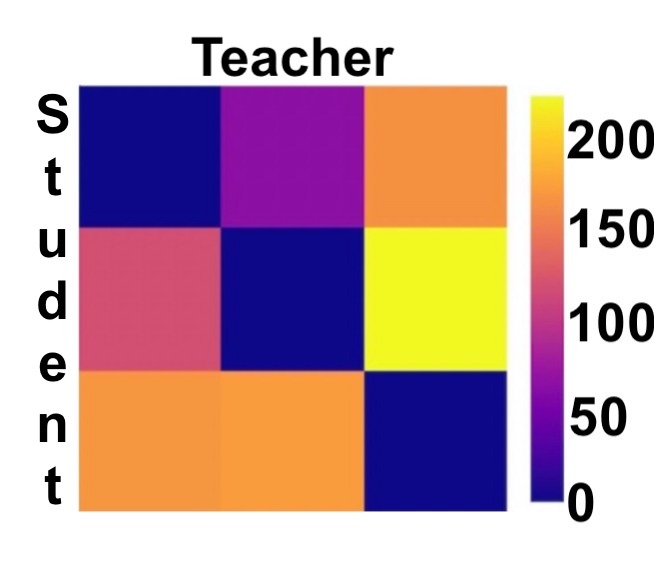}
        \caption{Experiment 4}
    \end{subfigure}
    \caption{Heatmap to visualize the number of counterfactuals distilled to each model. Rows are students and columns teachers, with yellow implying more instances and blue less.}
    \label{fig:matrix}
\end{figure}

\noindent
\textbf{Experiment 5: Many Model Distillation Between Semi-Experts.} We create ten datasets from the grayscale image dataset FashionMNIST~\cite{fashionMNIST}, each of which is undersampled (by 95\%) in \underline{one particular} (and different) class. This creates a scenario in which all models are deficient at predicting a particular class but other models are proficient in that class. This is a rigorous test of our claims of multi-way and selective distillation. To produce higher quality counterfactuals, we optimize images only over the 50\% most variable pixels of their class.

The median baseline accuracy for the ten models rests at 76\%. Since each model acts as a teacher to the other models, each model would receive thousands of new counterfactuals for the under-represented class, resulting in redundant counterfactuals which elevate performance to 79\%. After removing similar counterfactuals via geometric set-cover, we improve median accuracy of 83\%, approaching our topline (no undersampling) accuracy of 86\%. Since all models are networks of the same architecture, we could apply a greater range of competitors such as parameter transfer. This is the only experiment in which one of the competitors (knowledge distillation pretraining) surpasses our technique.

\noindent
\textbf{Experiment 6: Sensitivity to Feature Overlap.} Three random and non-overlapping subsets are extracted from the Statlog German Credit dataset with 400, 400, and 200 instances. We generate two models from the larger subsets and test on the third. Since we are using the same dataset, there is a perfect overlap between the features. Iteratively, we remove different features at random from both datasets until they only have two in common. This sensitivity test examines the effect of feature overlap on our method's performance and is repeated five times due to the random nature of feature removal. Since the datasets are random samples from the same distribution, performance increase is small compared to other algorithms but largely invariant to feature overlap, with correlation coefficients of $-9.45*10^{-4}$ and $1.17*10^{-3}$ for each model (Figure \ref{fig:AccOverOverlap}), averaging a very small ($2.25*10^{-4}$) correlation between increase in accuracy and difference in feature space. We speculate that perhaps this is due to the data distributions becoming more diverse as features are removed, allowing for more distillation between the models.  This indicates that data similarity may be an important indicator of success, and differences between datasets can overcome the feature overlap problem.

\noindent
\section{Understanding The Mechanisms of Distillation}
\label{sec:mech}
We test two  hypotheses to explain our previous successful experiments. First, we believe our method introduces novel virtual instances to a learners' dataset increasing diversity and allowing better generalization. To test this, we visualize the counterfactuals our method creates using t-SNE (Figure \ref{fig:exp1Distrib}) and find that counterfactuals, while being distinct from original data, create a similar distribution to it.

Our second hypothesis is that the teacher encodes information it believes to be important for classification into each instance. To test this hypothesis, we examine the average modification to original instances and separate them into counterfactuals of the positive ($\delta_+$) and negative ($\delta_-$) classes. Subtracting $\delta_-$ from $\delta_+$ provides a vector that details the changes made to features to move them from one class to another. Figure \ref{fig:exp4Heatmap} demonstrates such knowledge distilled for Experiment 4. Here, models receive different information learned from their teachers. For instance, models 2 and 3 are provided with the information that someone is more at risk for heart disease given an increase in age, while the counterfactuals generated for model 1 are produce relatively small changes to age. The diversity of these vectors illustrates how different models have discovered different patterns which may explain the models' performance increase.

The datasets of Experiment 2 have hundreds of features, making a visualization such as Figure \ref{fig:exp4Heatmap} uninterpretable. Instead, how features were to move the output closer to the 'expensive' class. Features that decreased the most were mileage (by an average of 3660 miles), years prior to 2005, and models such as Prius, Outback, and Range Rover Sport, and makes Toyota and Smart. Counterfactuals had the strongest positive association with years above 2013, models F-450 Super Duty, GX, and LX 570, and makes Masarati and Porsche. The CL dataset had few Toyota listings with no exposure to Porsche or LX 570 vehicles, indicating that the AucEx model successfully introduced such instances to the Cl model, partially explaining the performance increase.

\noindent

\section{Discussion \& Conclusion}

Our experiments demonstrate our approach can distill knowledge between two or more models regardless of architecture, algorithm, feature overlap, and under small or large data settings. Since our method targets specific weaknesses of each model, we can distill knowledge between any combination of high and/or low-performance models, compared to traditional knowledge distillation techniques which tend to only distill knowledge from a single high-performance model to a low-performance model~\cite{hinton}\cite{survey}. Though our method performed well on real-world data sets it does have some assumptions. It assumes there is some overlap between the features of the data sets and most importantly, our method works best when the distribution of the datasets used to train models are significantly different from each other. Further, our method is fundamentally limited by the strength of counterfactual generation. Counterfactual explanations are easy to compute on tabular data but their performance on more complex data, such as images is more challenging. However, more recent approaches have found success in more basic image networks~\cite{imageCF}~\cite{mnistCF}, so as research progresses, we believe this limitation will be removed.

We show in Figure \ref{fig:matrix} the number of instances each model teaches to the others. Interestingly, this quantity is asymmetrical which will motivate future work to better understand the mechanisms of how each model teaches the others.

\noindent
\textbf{Conclusion} We present a novel form of knowledge distillation that can be used between multiple models, in multiple directions and is focused. Each model simultaneously acts as teacher and student, distilling knowledge to the other by encoding learned information into virtual counterfactual instances and passing them into the training sets of other models. Unlike other knowledge distillation algorithms, which always distill knowledge from the teacher to student, we use a targeting mechanism to ensure that teachers only distill correct knowledge tailored to a student's deficiencies.

In our four main experiments, our method beats the competitors studied, including state-of-the-art knowledge distillation algorithms. In a stress test to determine if knowledge could be distilled between many (10) models, our model surpasses all but one competitor and remains competitive.
We find our method particularly useful in the setting where models can be freely shared, but raw data cannot, and the data sets share some features. This is common in medical imaging or finance communities where data is confidential. Given our method's strong performance on experiments simulating the aforementioned setting, we believe this to be a viable approach to knowledge distillation under such circumstances. 

\bibliography{ref}

\end{document}